\pdfoutput=1
\documentclass[11pt]{article}

\usepackage[preprint]{acl}

\usepackage{times}
\usepackage[T1]{fontenc}
\usepackage[utf8]{inputenc}
\usepackage{microtype}
\usepackage{inconsolata}
\usepackage{float}
\usepackage{booktabs}
\usepackage{amsmath}
\usepackage{newtxmath}
\usepackage{array}
\usepackage{xcolor}
\usepackage{graphicx}
\usepackage{tikz}
\usetikzlibrary{arrows.meta,positioning,fit,backgrounds,shapes.geometric}
\usepackage{pgfplots}
\pgfplotsset{compat=1.17}
\newif\ifprebuiltfigs
\prebuiltfigstrue
\usepackage{multirow}
\usepackage{algorithm}
\usepackage{algpseudocode}

\newcommand{\rupee}{\text{\small Rs.\,}}
\newcommand{\tb}{$\tau$-bench}
\newcommand{\ttb}{$\tau^2$-bench}

\newcommand{\bench}{TauIndianBankBench}
\newcommand{\Bench}{TauIndianBankBench}
\newcommand{\releasefootnote}{\raggedright Code:
\href{https://github.com/npci/TauIndianBankBench}{\nolinkurl{github.com/npci/TauIndianBankBench}};
dataset:
\href{https://huggingface.co/datasets/NPCI/tau-indian-banking}{\nolinkurl{huggingface.co/datasets/NPCI/tau-indian-banking}}.}

\title{TauIndianBankBench: Procedure-Aware, Action-Oriented\\ Evaluation of Tool-Using Conversational Agents}
\author{Krishanu Adhikary \qquad Prashant Devadiga \\
  National Payments Corporation of India (NPCI) \\
  Correspondence: \texttt{krishanu.adhikary@npci.org.in}}

\begin{document}
\maketitle

\begin{abstract}
Many Indian banks serve customers through FAQ pages and scripted chatbots.
Assistants based on language models could respond faster and carry out requests
such as blocking a card, but only if they give correct information and follow
the bank's rules. These rules prescribe a \emph{procedure}, a set of required
steps, such as obtaining a closure quote before closing a deposit. Adapting
models to this domain requires measuring how well they follow the procedure,
before and after training. Benchmarks usually judge only the records left at
the end of a conversation. Many required steps, however, leave the records
unchanged, so an assistant that skips them can still be judged successful. We
present \bench{}, an action-oriented benchmark of 1{,}000 Indian retail-banking
tasks where an assistant helps a simulated customer. Besides the final records,
it checks that the assistant took every required step and made no unrequested
change. For every model, 6.5--12.0\% of sessions are judged successful when
only the final records are checked, although the assistant skipped a required
step or made an unrequested change. Scoring both procedure and outcome shows
more accurately how well language models perform in Indian retail banking.

\end{abstract}

\section{Introduction}
\label{sec:intro}

Many Indian banks serve customers through FAQ pages and scripted chatbots that
return fixed answers. Assistants based on language models could respond faster, answer
questions that no FAQ anticipates, and carry out requests such as blocking a
card or booking a deposit. A bank can rely on such an assistant only if it gives
correct information and follows the bank's rules, because an error affects the
customer directly. Adapting a model to this domain, for example by training it,
therefore requires a benchmark that measures how well the model serves
customers, both before and after training.

Several benchmarks evaluate language models as customer-service agents that
use tools to retrieve records, apply written policy, and make changes
\citep{yao2024taubench,barres2025tau2,huang2024crmarena}. These
benchmarks usually check the resulting records rather than the transcript,
because an assistant may describe blocking a compromised card in fluent prose
without calling the tool that blocks it, and a metric that can be satisfied
without doing the work can be gamed \citep{skalse2022reward,zhu2025rigorous}. Checking the final
records has its own limitation, however: written policy prescribes a
\emph{procedure} of required steps, and many of these steps leave the records
unchanged.

Before closing a deposit, the assistant must obtain a closure quote so that the
customer knows the penalty (Figure~\ref{fig:order}). Before booking a deposit
that the balance may not support, it must look up the bank's minimum-balance
rule. Before blocking a card, it must check the card's status. An assistant
that skips these steps still produces the correct
records and passes a final-state check, but the customer does not receive the
information that these steps are meant to provide.

In Indian retail banking, regulation states many required steps explicitly:
know-your-customer (KYC) verification must be renewed periodically \citep{rbi2016kyc}, deposit interest is subject to tax
deducted at source (TDS) with prescribed exemption forms, and government
savings schemes have statutory eligibility rules. Our tasks turn the bank's
written policy for servicing requests, such as closing a deposit or applying
a senior-citizen rate, into required steps, most of which are
reads. In our corpus, 962 of the 1{,}000 released tasks require at least one
read-only call, and 114 require no write call. Existing Indian and financial
evaluation suites assess language coverage, domain knowledge, numerical
reasoning, or financial tool use, but not customer servicing under Indian
banking rules (\S\ref{sec:related}).

We therefore introduce \bench{},\footnote{\releasefootnote} an
action-oriented benchmark of Indian retail-banking servicing
(\S\ref{sec:environment}). It comprises a synthetic bank, a catalogue of
servicing tools, a written policy, a knowledge base that covers these
regulations, and 1{,}000 tasks, from which smaller train and test splits are
drawn for adapting a model to the domain. It runs in the \ttb{}
harness \citep{barres2025tau2}, an extension of \tb{} \citep{yao2024taubench}.
In this harness, the assistant under evaluation talks with a simulated
customer, itself a language model, and acts on a shared database. Tasks fall
into four categories. \emph{Sequencing} tasks require several dependent calls
(up to six). \emph{Edge-case} tasks contain a condition that the assistant must
recognise before acting, such as an ineligible or ambiguous request.
\emph{Rare-tool} tasks use infrequently needed operations. \emph{Control} tasks
are straightforward requests and serve as an internal baseline.

Each session receives two pass/fail verdicts (\S\ref{sec:scoring}). The
\emph{final-state check}, inherited from \tb{} and \ttb{}, compares the final records
with those produced by replaying the task's \emph{gold chain}, its reference
sequence of tool calls. The \emph{action check} extends the action matching of
\ttb{}. It requires every call in the gold chain, including reads, to appear in
the session. Its \emph{write purity} condition fails the session if the
assistant makes any change that the task did not request. Our headline
measure is the pass rate, the share of sessions judged successful; a session
passes only if both checks pass.

We evaluate six open-weight models, five from the Gemma~4 family
\citep{gemma2026gemma4} and Qwen3.8-27B \citep{qwen2026qwen38}, with the same
simulated customer, and report two findings (\S\ref{sec:results}). First,
the final-state check accepts procedural violations for every model: it passes
sessions in which the assistant omitted a required call or made an unrequested
change (\S\ref{sec:exp1}). Second, each check fails sessions that the other
passes. For the three strongest models, most sessions that fail exactly one
check are procedural violations, which only the action check detects. For the
weakest model, most such sessions include every gold-chain call yet end with
incorrect records, which only the final-state check detects
(\S\ref{sec:exp2}). Two further results support the
validity of this pass criterion. Trivial scripted agents pass no task
(\S\ref{sec:exp7}), and control-task pass rates show no trend with model size,
whereas the three graded categories separate the models (\S\ref{sec:exp5}).

Our contributions are as follows: (1)~an Indian retail-banking environment
with a synthetic bank, servicing tools, and tasks derived from written policy
(\S\ref{sec:environment});
(2)~an action check that complements the final-state check by scoring required
reads and penalising unrequested changes (\S\ref{sec:scoring}); and (3)~an
evaluation of six models under both checks, showing that the final-state check
alone gives an incomplete measure of how well models follow the procedure (\S\ref{sec:results}). We release the
code, data, and reference results.

\section{Related Work}
\label{sec:related}

\paragraph{The \tb{} family.}
\Bench{} is a domain of the \tb{}/\ttb{} family
\citep{yao2024taubench,barres2025tau2}. In \tb{}, a task is passed when the
final database matches the expected one and the agent has conveyed the
required information to the user \citep{yao2024taubench}. \ttb{} adds dual control, in which the simulated user also acts on shared
state. It also adds \emph{action matching}, which checks that each call of the
reference solution appears in the session \citep{barres2025tau2}.
$\tau$-Knowledge adds $\tau$-Banking, a customer-support domain in which the
agent searches about 700 policy documents while updating accounts. Success is
judged on the final database and, for some tasks, on action matching
\citep{shi2026tauknowledge}. FraudBench builds
adversarial callers on the same banking environment
\citep{pai2026fraudbench}. \Bench{} builds on these designs: it requires every
read that its written policy mandates, and it fails a session in which the
assistant makes an unrequested change and later reverts it.

\paragraph{Conversational and enterprise tool-use benchmarks.}
Function-calling benchmarks evaluate API selection and arguments, from ToolLLM's
more than 16{,}000 real APIs \citep{qin2023toolllm} to the Berkeley Function
Calling Leaderboard \citep{patil2025bfcl} and ACEBench
\citep{chen2025acebench}. AgentBench evaluates models as agents across eight
environments \citep{liu2023agentbench}, API-Bank evaluates API planning and
calling in annotated dialogues \citep{li2023apibank}, and ToolTalk specifies
multi-step intents through dialogue \citep{farn2023tooltalk}. Interactive and
enterprise benchmarks execute the agent and assess the outcome: WebArena and
Mind2Web on web tasks \citep{zhou2024webarena,deng2023mind2web}, WorkArena on
ServiceNow knowledge work \citep{drouin2024workarena}, and CRMArena and
CRMArena-Pro on customer-relationship tasks, the latter with persona-driven
users \citep{huang2024crmarena,huang2025crmarenapro}. AppWorld additionally
checks for collateral damage, that is, unrequested changes
\citep{trivedi2024appworld}, and Thinkingbox rejects wrong, missing, or extra
effects in the terminal state of business workflows \citep{li2026thinkingbox}.
These interactive benchmarks assess what the agent changed; they cannot detect a required step that modifies no
records.

\paragraph{Process and policy-compliance scoring.}
Other work scores the individual steps of an agent. TRAJECT-Bench checks tool
choice, arguments, and dependency order against reference trajectories
\citep{he2025trajectbench}, TRACE scores reasoning steps without a reference
\citep{kim2025trace}, and EgoBench validates both process and result
\citep{liu2026egobench}. For mathematical reasoning, reward models trained with
step-level supervision outperform those trained only on final answers
\citep{lightman2023verify}. ST-WebAgentBench credits only policy-compliant
completions \citep{levy2024stwebagentbench}, SOP-Bench builds tasks from
expert-written standard operating procedures \citep{nandi2025sopbench}, and
IntellAgent generates test conversations from a policy graph
\citep{levi2025intellagent}. ToolSandbox is the closest design \citep{lu2024toolsandbox}. It checks
required \emph{milestones}, including specific tool calls, and forbidden
\emph{minefields}. Its milestone score, however, gives partial credit for
similarity to the expected session, including message text, and minefields must
be written for each task. JourneyBench
\citep{balaji2026journeybench} scores a conversation zero if any tool call is
missing, extra, or out of order, but its tools return pre-generated responses,
so it cannot check whether the calls produced the correct records. Our benchmark combines a shared database and a final-state check
with an action check that requires policy-mandated reads and enforces write
purity (\S\ref{sec:scoring}). We apply both checks to every session and measure
the gap between them (\S\ref{sec:results}).
Table~\ref{tab:taucompare} (Appendix~\ref{app:compare}) summarises these
designs.

\paragraph{Financial and Indian-language benchmarks.}
Most financial LLM benchmarks test knowledge and numerical reasoning: broad
suites such as FinBen, PIXIU, and FinEval
\citep{xie2024finben,xie2023pixiu,guo2023fineval}, numerical and table
reasoning \citep{chen2021finqa,chen2022convfinqa,zhu2021tatqa,krumdick2024bizbench},
and deposit and loan arithmetic \citep{lee2026bankmathbench}. The Finance Agent
Benchmark, FinToolBench, and FinMCP-Bench let the model call tools
\citep{bigeard2025financeagent,lu2026fintoolbench,zhu2026finmcpbench}. Evaluation in the Indian context has focused
on Indian languages
\citep{kakwani2020indicnlpsuite,kumar2022indicnlg,doddapaneni2023indicxtreme}
and India-specific knowledge \citep{verma2024milu}, with a bilingual finance
subset in BhashaBench \citep{devane2025bhashabench}.

\paragraph{Validity of agent evaluation.}
The agentic benchmark checklist of \citet{zhu2025rigorous} recommends
trivial-agent baselines to expose scores that can be
obtained without performing the task. An audit of tool-calling benchmarks identifies brittle state matching and trajectory lock-in
\citep{bhat2026validity}. LLM-simulated users can misestimate agent success
\citep{seshadri2026lostinsimulation}, and agents degrade when users do not
cooperate \citep{shim2025noncollab}. Judge models exhibit position, verbosity,
and self-preference biases \citep{zheng2023mtbench,liu2023geval}, which
motivates deterministic scoring.

\section{Benchmark Design}

\subsection{Environment and tasks}
\label{sec:environment}

\paragraph{Simulated bank.}
A \emph{session} is one conversation on one task between the \emph{assistant}
(the agent under evaluation) and the simulated customer. Each session runs on
a fresh copy of a synthetic bank of 944 customers, each with accounts, cards,
deposits, loans, standing instructions, a KYC status, and a transaction
history. The clock is fixed and generated identifiers are seeded, so the
replay of the gold chain used for scoring (\S\ref{sec:scoring}) starts from the
same bank state as the session. The assistant receives the bank's written policy and can search, by keyword,
24 knowledge-base documents on payment limits, KYC renewal, deposit insurance,
tax on interest, chargeback timelines, and government schemes. It
has access to 34 tools: 33 servicing tools for accounts, deposits, loans,
cards, mandates (standing payment authorisations), cheques, and service
requests, and one that transfers the customer to a human staff member.
Thirteen servicing tools are \emph{writes}. Eleven of them change records; the
other two request a cheque book or a duplicate statement without changing any
record, and the scorer treats them as writes like the rest. The remaining
twenty servicing tools only retrieve data (\emph{reads};
Table~\ref{tab:catalog}, Appendix~\ref{app:catalog}).

\paragraph{Task specification.}
Each task fixes the session date and names the logged-in customer but
withholds the customer's profile, so the assistant must retrieve accounts,
cards, and products through the tools. Each task also specifies the simulated
customer's persona and scenario. Finally, each task specifies its \emph{gold chain}: the reference
sequence of tool calls that resolves the request, together with the arguments
each call must match. Gold chains contain 3.54 calls on average and range from
one to six (per-length counts in Table~\ref{tab:exp3},
Appendix~\ref{app:len}). Appendix~\ref{app:example} reproduces one released task.

\paragraph{Task categories and service areas.}
The tasks are drawn from 50 \emph{task families}, groups of tasks that share a
request type (listed in Table~\ref{tab:families}). The families fall into four categories
(Table~\ref{tab:categories}, Appendix~\ref{app:catstats}): three
\emph{graded} categories designed to be difficult, and a \emph{control}
category of simple requests that serves as a baseline. By subject, the tasks fall into four service areas: everyday
banking (balances, mandates, cheques, service requests), deposits, loans
(instalments, foreclosure quotes, gold loans), and cards (controls, blocking, disputes).

\paragraph{Data splits.}
The release defines four splits: \emph{base} (1{,}000 tasks), on which all
reference results are computed; \emph{lite}, a subset of 200 tasks, four
from each family, for faster evaluation; and \emph{train} (150) and \emph{test} (50),
which partition lite. Scores from different splits are not comparable, and a
method tuned on the train split should report its results on the test split.

\paragraph{Task construction and validation.}
A language model drafted the tasks against the live environment, and a task
was admitted only if it passed four filters. The first filter enforces varied first messages. Within a family, no first
message may occur in more than 40\% of the tasks, and at least eight distinct
ones are required. This prevents a model from identifying the family instead of
interpreting the request. The second requires a gold chain of at least three calls for graded tasks. The third ensures
that no task appears in both the train and test splits. The fourth admits a
task only if its replayed gold chain passes both checks (\S\ref{sec:scoring}).

\subsection{Scoring}
\label{sec:scoring}

A session is scored deterministically by two pass/fail checks. The
\emph{action check} examines the \emph{trajectory}, the record of every
message and tool call; the \emph{final-state check} examines the final
database. The final-state check is inherited from \tb{} and \ttb{}
\citep{yao2024taubench,barres2025tau2}, and the requirement that each
gold-chain call appear in the session comes from the action matching of \ttb{}. We add two components
(\S\ref{sec:related}): the gold chain includes the reads that the policy
requires, and \emph{write purity} requires every write to match a gold-chain
write.

\paragraph{Pass criterion.}
A session scores $1$ if both checks pass and $0$ otherwise, and a session that
scores $1$ \emph{passes}. The pass rate is the headline score of the release
(column All of Table~\ref{tab:ladder}); unless stated otherwise, every score in
\S\ref{sec:results} is a pass rate. A session thus passes exactly when it ends
normally (\S\ref{sec:protocol}), the assistant makes every gold-chain call and
no unrequested change, and the final records match the expected ones. No partial credit is given: a session that makes three of four required calls
scores $0$, the same as one that does nothing. Appendix~\ref{app:pseudocode} gives the
pseudocode.

\paragraph{Final-state check.}
\emph{Oracle replay} runs the task's gold chain on the same seeded starting
records to produce the expected final records, and the final-state check
compares the session's final records with them. Before the comparison, the
scorer normalises both: it removes fields that change from run to run
(generated references, download links, free-text notes) and ignores the
position of entries in collections keyed by generated identifiers. On service
requests, it also ignores the optional account and treats the two general
request categories as one. Timestamps need no removal, because the environment
runs on a fixed clock. The scorer then compares hashes of the normalised
records, so records match when their content is identical, regardless of their
byte-level representation. The check therefore observes
only the outcome: because reads modify no records, it cannot detect an omitted
read, and on a task that requires no change, any session that makes no change
passes it. The action check addresses this limitation.

\begin{figure}[t]
\centering
\ifprebuiltfigs
\includegraphics[width=\columnwidth]{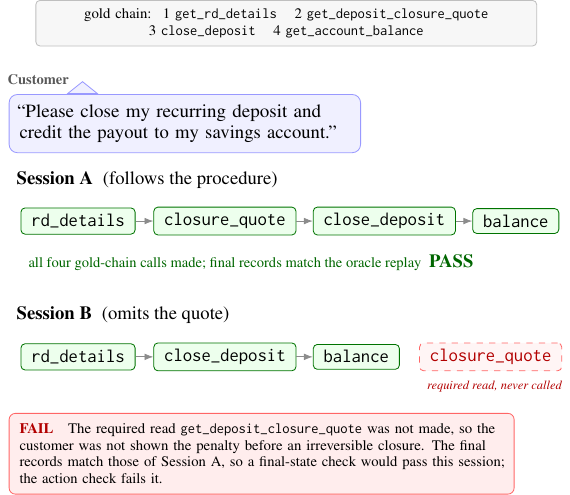}
\else
\resizebox{\columnwidth}{!}{%
\begin{tikzpicture}[
  font=\small,
  cb/.style={draw=blue!40, fill=blue!6, rounded corners=4pt, align=left,
             inner sep=5pt, text width=56mm},
  chip/.style={draw=green!45!black, fill=green!7, rounded corners=2pt,
               font=\footnotesize\ttfamily, inner xsep=5pt, inner ysep=3pt},
  ghost/.style={draw=red!60, dashed, fill=red!4, rounded corners=2pt,
               font=\footnotesize\ttfamily, inner xsep=5pt, inner ysep=3pt, text=red!70!black},
  arr/.style={-{Latex[length=1.5mm]}, black!45},
  who/.style={font=\scriptsize\bfseries, text=black!65},
  sess/.style={font=\small\bfseries}]

% shared gold chain
\node[draw=black!25, fill=black!3, rounded corners=2pt, font=\scriptsize,
      inner sep=4pt, text width=82mm, align=center] at (4.3,0)
 {gold chain:\; 1 \texttt{get\_rd\_details} \;\; 2 \texttt{get\_deposit\_closure\_quote}\\
  3 \texttt{close\_deposit} \;\; 4 \texttt{get\_account\_balance}};

% shared customer request
\node[who] at (0.1,-0.95) {Customer};
\node[cb, anchor=north west] at (-0.4,-1.2)
 {``Please close my recurring deposit and credit the payout to my savings account.''};
\draw[fill=blue!6, draw=blue!40] (0.6,-1.19) -- (1.1,-1.19) -- (0.85,-0.99) -- cycle;

% ---------------- Session A ----------------
\node[sess, anchor=west] at (-0.4,-2.65) {Session A\; \normalfont\small (follows the procedure)};
\node[chip, anchor=west] (a1) at (-0.2,-3.35) {rd\_details};
\node[chip, anchor=west] (a2) at (2.05,-3.35) {closure\_quote};
\node[chip, anchor=west] (a3) at (4.75,-3.35) {close\_deposit};
\node[chip, anchor=west] (a4) at (7.45,-3.35) {balance};
\draw[arr] (a1)--(a2); \draw[arr] (a2)--(a3); \draw[arr] (a3)--(a4);
\node[font=\scriptsize, text=green!40!black, anchor=west] at (-0.2,-4.05)
 {all four gold-chain calls made; final records match the oracle replay
  \;{\small\textbf{PASS}}};

% ---------------- Session B ----------------
\node[sess, anchor=west] at (-0.4,-4.95) {Session B\; \normalfont\small (omits the quote)};
\node[chip, anchor=west] (b1) at (-0.2,-5.65) {rd\_details};
\node[chip, anchor=west] (b3) at (2.05,-5.65) {close\_deposit};
\node[chip, anchor=west] (b4) at (4.75,-5.65) {balance};
\draw[arr] (b1)--(b3); \draw[arr] (b3)--(b4);
\node[ghost, anchor=west] (bg) at (6.55,-5.65) {closure\_quote};
\node[font=\tiny\itshape, text=red!70!black, anchor=west] at (6.55,-6.15) {required read, never called};

% why it fails
\node[draw=red!60, fill=red!7, rounded corners=3pt, align=left, inner sep=5pt,
      text width=82mm, font=\scriptsize, anchor=north west] at (-0.4,-6.6)
 {\textbf{\textcolor{red!70!black}{FAIL}}\quad The required read
  \texttt{get\_deposit\_closure\_quote} was not made, so the customer was not
  shown the penalty before an irreversible closure. The final records match
  those of Session~A, so a final-state check would pass this session; the
  action check fails it.};

\end{tikzpicture}}
\fi
\caption{One task, two sessions. The customer asks the assistant to close a
recurring deposit. Session~A obtains the closure quote that the task requires and
passes; Session~B omits it and fails, although both sessions leave identical
records (a related task appears in Appendix~\ref{app:example}).}
\label{fig:order}
\end{figure}
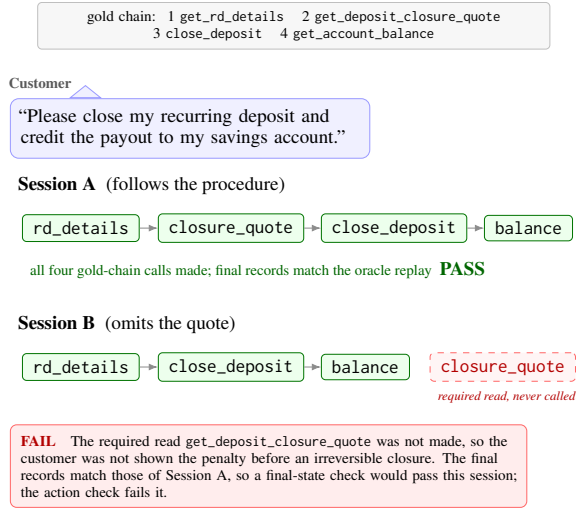

\paragraph{Action check.}
The action check searches the session for each gold-chain call, regardless of
the order in which the calls were made. Each gold-chain call specifies which of
its arguments must match. If it lists some arguments, only those are compared.
If its argument list is empty, only the tool name is compared. If it has no
argument list, every argument given in the gold-chain call is compared, except
run-dependent identifiers and free-text fields, which carry no information
about correctness. An optional account identifier is compared only when both
the gold-chain call and the assistant supply it. A call that appears twice in
the gold chain is satisfied by a single matching call. Because the gold chain
includes the required reads, a session that omits one fails the action check
even when its final records are correct (Figure~\ref{fig:order}).

\paragraph{Write purity.}
If the assistant closes the wrong deposit or freezes a card the customer
did not mention, write purity fails the session even if every gold-chain call
was also made. This prevents an assistant from passing by issuing every
available call. The rule also applies when the change is later undone: the
final-state check then passes, but write purity fails the session, because the
customer was affected while the change was in place.

\paragraph{Call order.}
The task design fixes the order of the required steps in two ways (column
Order of Table~\ref{tab:taucompare}). First, some calls need values that only an earlier call returns: in
the task of Appendix~\ref{app:example}, the maturity projection needs the rate
that the rate-card call returns, so it can match the gold chain only after
that call. Second, in every task outside the control category, the simulated
customer's instructions list the requests as numbered steps, which the customer
raises one at a time and in order. In the same task, for instance, the customer
agrees to book the deposit only after seeing the projected maturity value. The task design thus
determines when each step can occur, the action check verifies that each
required step occurred, and the final-state check verifies the result.

\paragraph{Scorer audit.}
We compared the released scorer with an independent re-implementation of the
scorer on 600 recorded sessions; the two implementations gave the same
verdicts on every session and on all $705$ paired tool calls. Before release,
the packaged scorer also re-scored 5{,}000 recorded sessions and reproduced
every score exactly. The model ranking (\S\ref{sec:results}) was also reproduced on a
second model-serving setup, with score differences no larger than those
between repeated runs. The harness's judge model, which grades the conversation
text and does not affect the released score, gives the same verdict as the
re-implementation on $87.5\%$ of sessions and gives less credit in every
disagreement (Table~\ref{tab:audit}, Appendix~\ref{app:audit}).

\section{Evaluation Protocol}
\label{sec:protocol}

\paragraph{Conversation protocol.}
A session alternates between the assistant and the customer. On each turn,
the assistant either sends one message to the customer or makes one or more
tool calls, but not both. The session ends when the customer states that the request is resolved, or
when the assistant ends the session or transfers the customer to a human staff member. A
session that reaches the limit of 200 steps or 10 failed tool calls, or stops
because of an error, scores $0$. A
language model plays the customer. It is prompted with the task's persona and
scenario and a fixed set of instructions. It reveals information step by step
and does not name tools or say which call to make. In graded tasks, it asks for
the steps in the listed order and asks the assistant to return to a step it
skipped.

\paragraph{Reporting.}
Because the customer simulator shapes every conversation, we report the
assistant model, the customer-simulator model, and the split together, and
compare only runs that share all three. Throughout this paper, MiniMax-M2.7
\citep{minimax2026m2} plays the customer at temperature zero, so that the
customer's side of every conversation can be reproduced. A run may attempt
each task $k$ times to estimate $\text{pass}^k$, the probability that all $k$
attempts at a task pass. The reference results use one attempt per task on
the full base split, and Appendix~\ref{app:band} analyses a second attempt.

\section{Results}
\label{sec:results}

We evaluate five Gemma~4 models \citep{gemma2026gemma4}, namely the dense E2B,
E4B, 12B and 31B models and the 26B-A4B mixture-of-experts model, together with
Qwen3.8-27B \citep{qwen2026qwen38}. All results are computed on the
1{,}000-task base split, with one attempt per task under each model's most deterministic sampling setting and the same
simulated customer for all models (\S\ref{sec:protocol}). The 95\% Wilson
confidence interval of a pass rate is approximately $\pm3$ points over the full
split and at most $\pm8$ points within a single category. Each analysis that
compares scoring rules re-scores the same recorded sessions, so that any
difference between two scores reflects the scoring rule rather than a new run.

\subsection{Reference results}
\label{sec:ladder}

\begin{table}[t]
\centering\small\setlength{\tabcolsep}{3pt}
\begin{tabular}{@{}lrrrrr@{}}
\toprule
Assistant model & All & Seq & Edge & Tools & Ctrl \\
\midrule
Qwen3.8-27B      & 0.849 & 0.921 & 0.774 & 0.757 & 0.848 \\
Gemma\,4 31B     & 0.776 & 0.868 & 0.718 & 0.559 & 0.821 \\
Gemma\,4 26B-A4B & 0.714 & 0.802 & 0.662 & 0.500 & 0.759 \\
Gemma\,4 12B     & 0.699 & 0.732 & 0.639 & 0.533 & 0.929 \\
Gemma\,4 E4B     & 0.607 & 0.662 & 0.500 & 0.480 & 0.804 \\
Gemma\,4 E2B     & 0.479 & 0.455 & 0.426 & 0.388 & 0.830 \\
\bottomrule
\end{tabular}
\caption{Reference results: pass rate on the 1{,}000-task base split,
overall (\textbf{All}) and by category (\textbf{Seq}=sequencing,
\textbf{Edge}=edge case, \textbf{Tools}=rare tool, \textbf{Ctrl}=control).
Qwen3.8-27B and the 26B-A4B mixture-of-experts model are shown for comparison
with the four dense Gemma models.}
\label{tab:ladder}
\end{table}

Table~\ref{tab:ladder} reports the reference results. Among the four dense Gemma models, each larger model passes more tasks than the
next smaller one ($p<0.005$ for each pair, paired sign test). This ranking
also holds within each of the three graded categories (Figure~\ref{fig:cat},
Appendix~\ref{app:exp34}).

\subsection{Violations missed by the final-state check}
\label{sec:exp1}
We score every session twice: with the final-state check alone, as in \tb{},
and with both checks, as in the headline score. A
session that passes the former but fails the latter ended with the correct
records but either omitted a required call or made an unrequested change; we
refer to such a session as a \emph{procedural violation}
(Table~\ref{tab:exp1}; Figure~\ref{fig:inflate}, Appendix~\ref{app:exp34}).

\begin{table}[t]
\centering\small\setlength{\tabcolsep}{3pt}
\begin{tabular}{@{}lrrrl@{}}
\toprule
Assistant model & \shortstack[r]{Final-state\\only} & $b$ & $c$ & McNemar $p$ \\
\midrule
Qwen3.8-27B      & 0.920 & 71  & 13  & $<10^{-10}$ \\
Gemma\,4 31B     & 0.896 & 120 & 25  & $<10^{-15}$ \\
Gemma\,4 26B-A4B & 0.821 & 107 & 37  & $<10^{-8}$ \\
Gemma\,4 12B     & 0.764 & 65  & 88  & $0.075$ (n.s.) \\
Gemma\,4 E4B     & 0.725 & 118 & 109 & $0.596$ (n.s.) \\
Gemma\,4 E2B     & 0.598 & 118 & 180 & $<10^{-3}$ \\
\bottomrule
\end{tabular}
\caption{Disagreements between the two checks on the sessions of
Table~\ref{tab:ladder}. \textbf{Final-state only}: pass rate under the
final-state check alone. $b$: sessions that ended with the correct records but
failed the action check (procedural violations). $c$: sessions that passed the
action check but ended with incorrect records. McNemar $p$: two-sided exact
test of whether $b$ and $c$ differ.}
\label{tab:exp1}
\end{table}

Scoring only the final records raises each model's pass rate by $6.5$--$12.0$
points, the share of its sessions that are procedural violations. Table~\ref{tab:causes} decomposes action-check failures by
cause over all failed sessions, not only these violations. A missing required write is the most frequent cause for four of the six
models, and a missing required read for the two smallest (E4B and E2B).
Unrequested writes are the least frequent cause for every model except
Qwen3.8-27B.
The violation rate is highest in loan tasks, because no loan tool is a write and
the final-state check therefore cannot observe an omitted loan step
(Appendix~\ref{sec:exp6}).

\subsection{Complementarity of the two checks}
\label{sec:exp2}
The two checks disagree only on the $b$ and $c$ sessions of
Table~\ref{tab:exp1}. For the three strongest models (Qwen3.8-27B, Gemma\,4
31B and 26B-A4B), $b$ exceeds $c$ (all $p<10^{-8}$). For these models, most sessions that fail only one check are
procedural violations, which only the action check detects. For the weakest
model, E2B, the reverse holds ($p<10^{-3}$): most such
sessions include every gold-chain call yet end with incorrect records, an error
that only the final-state check detects. For the 12B and E4B, $b$ and $c$ do
not differ significantly. Neither check can therefore substitute for the other.

\subsection{Performance by task category}
\label{sec:categories}
The categories differ in what they require and in how widely they separate the
models; Table~\ref{tab:catstats} (Appendix~\ref{app:catstats}) summarises their
composition.

\paragraph{Sequencing.} Sequencing tasks (470 tasks in 20 families; Table~\ref{tab:families}) require
several dependent calls. Although their gold chains are the shortest
among the graded categories ($3.64$ calls on average; 279 tasks require three
calls), this category separates the models most widely, from $0.455$ for E2B to $0.921$ for Qwen3.8-27B.

\paragraph{Edge cases.} Edge-case tasks (266 tasks in 14 families; Table~\ref{tab:families}) each contain a condition that the
assistant must recognise before acting (Appendix~\ref{app:catstats} groups
these families). Every edge-case gold chain
contains at least one write, so the assistant must both handle the exception
and complete the legitimate part of the request. Every model scores lower on
edge cases than on sequencing, with pass rates from $0.426$ (E2B) to $0.774$
(Qwen3.8-27B).

\paragraph{Rare tools.} Rare-tool tasks (152 tasks in 8 families; Table~\ref{tab:families}) are each built around an
infrequently required operation, such as a deposit maturity projection, the
status of an earlier service request, or the details of an insurance policy.
None of these eight tools appears in more than 56 of the 1{,}000 gold chains. This is the lowest-scoring category for every model. The
Gemma models fall within a narrow range ($0.388$--$0.559$), whereas Qwen3.8-27B
exceeds the 31B by $0.198$, its largest margin over that model in any category.

\subsection{Effect of procedure length}
\label{sec:exp3}
Tasks that require more calls are not uniformly harder: four-call tasks are the
hardest for every model except E2B. The gap between the strongest and the
weakest model is widest on five-call tasks ($0.614$, against $0.024$ on
one-call tasks). Within the sequencing and edge-case categories, long tasks affect the weakest
models most. For example, E4B falls from $0.810$ on short sequencing tasks to
$0.445$ on long ones, whereas none of the three strongest models changes by
more than $0.086$ on sequencing or edge-case tasks. On rare-tool
tasks, by contrast, Qwen3.8-27B declines most, from $0.923$ to $0.699$
(Appendix~\ref{app:len}).

\subsection{Control category}
\label{sec:exp5}
The control category (112 tasks, $1.38$ calls on average, 93 with no write)
allows us to exclude an alternative explanation, namely that the graded
categories reward general capability rather than competence in banking
procedure. Under that explanation, pass rates on the simple control requests
would also increase with model size. Across the four dense Gemma models,
however, they remain between $0.804$ and $0.929$ without a trend in size, and
the highest rate belongs to the 12B, whereas sequencing pass rates increase
with size (\S\ref{sec:ladder}; Figure~\ref{fig:cat}). The categories that require multi-step
procedures thus rank the models by size, whereas the control category does
not.

\subsection{Trivial-agent baselines}
\label{sec:exp7}
We evaluated four scripted agents, none of which passes any task. The first
only replies to the customer, the second looks up records but makes no required
call, and the third claims in a persuasive message that the work is complete;
all three fail the action check. The fourth makes every available change and
fails write purity.

\subsection{Further analyses}
\label{sec:further}
Appendix~\ref{sec:exp4} classifies failed sessions by type. Stopping short of
the full procedure is the most frequent type for every model. Unrequested
writes make up a larger share of failures for the strongest models, and
sessions that finish with errors make up a larger share for the weakest. A
second run of the five Gemma models changes each pass rate by at most $2.7$ points, although 14--26\% of tasks change
outcome between the two runs (Appendix~\ref{app:band}).

\section{Conclusion}
Assistants based on language models could serve the customers of Indian banks,
but only if they follow the bank's procedure as well as reach the correct
result. Scoring only the final records gives an incomplete measure of this,
because many required steps do not change the records. \Bench{} adds an
action check that scores these steps. The final-state check alone accepts
procedural violations for every model, and each check fails sessions that the
other passes. We therefore recommend that benchmarks for service assistants in
regulated domains report an action check alongside the final-state check, and
that a model adapted to Indian retail banking be measured with both checks
before and after training, for which the release provides separate train and
test splits. The benchmark is distributed as a standard domain of the \ttb{}
harness \citep{barres2025tau2} and runs against any OpenAI-compatible
endpoint.

\section*{Limitations}
First, a score depends on both the assistant and the simulated customer
(\S\ref{sec:protocol}); the reported numbers therefore hold only for the
simulator we used, whose own error rate we did not measure. Second, all customer data are synthetic, so the
frequency of each type of edge case reflects our generated records rather than
actual bank traffic. Third, version~1.0.0 covers Indian retail banking in
English only; other languages and domains are left to future versions.
Finally, evaluation
is computationally costly. Every session is a live multi-turn conversation in which both the assistant and
the simulated customer are language models. A single run on the base split
therefore requires 1{,}000 such conversations of up to 200 steps each, and
estimating $\text{pass}^k$ multiplies this cost by $k$. When either model is
served through a commercial API, a full evaluation incurs substantial inference
cost. The 200-task lite split is provided for development and ablations, and
the base split for final reporting of untuned models.

\section*{Ethics Statement}
All customer data in the benchmark are synthetic, and the records contain no
field for a national identification number. Email addresses and phone
numbers are generated placeholders; all organisation and place
names are invented, and we verified that none matches a real one. The
benchmark is intended for research on and evaluation of tool-using agents and
does not constitute financial advice. A pass indicates only that, with our simulated customer, the assistant made the
required calls, made no unrequested change, and left the records correct. It
does not establish that the assistant is suitable for deployment in a bank,
which would require further safety, privacy, and regulatory review.

\bibliography{custom}

\appendix
\raggedbottom
\section{Comparison with Related Benchmarks}
\label{app:compare}
\begin{table}[H]
\centering\footnotesize\setlength{\tabcolsep}{2.6pt}
\begin{tabular}{@{}lccccc@{}}
\toprule
Benchmark & State & Set & Reads & Purity & Order \\
\midrule
\tb{}            & \checkmark & --         & --         & $\sim$     & -- \\
\ttb{}           & \checkmark & $\sim$     & --         & $\sim$     & -- \\
$\tau$-Knowledge & \checkmark & $\sim$     & $\sim$     & $\sim$     & -- \\
ToolSandbox      & $\sim$     & $\sim$     & $\sim$     & $\sim$     & -- \\
AppWorld         & \checkmark & --         & --         & $\sim$     & -- \\
JourneyBench     & --         & \checkmark & \checkmark & \checkmark & \checkmark \\
TRAJECT-Bench    & --         & \checkmark & \checkmark & $\sim$     & $\sim$ \\
Ours             & \checkmark & \checkmark & \checkmark & \checkmark & \checkmark \\
\bottomrule
\end{tabular}
\caption{What each benchmark scores and whether its tasks depend on call order: \tb{} \citep{yao2024taubench}, \ttb{}
\citep{barres2025tau2}, $\tau$-Knowledge \citep{shi2026tauknowledge},
ToolSandbox \citep{lu2024toolsandbox}, AppWorld \citep{trivedi2024appworld},
JourneyBench \citep{balaji2026journeybench}, TRAJECT-Bench
\citep{he2025trajectbench}, and ours. \checkmark{}: fully; $\sim$: partly or
with partial credit; --: not at all. \textbf{State}: final records compared
with a reference. \textbf{Set}: presence of the required calls, in any order,
required for a pass. \textbf{Reads}: required read-only calls checked.
\textbf{Purity}: an unrequested change fails the session (in \tb{}, \ttb{}
and $\tau$-Knowledge, through the final-state comparison). \textbf{Order}: call order matters for a pass, because the scorer checks it or, in ours, because the task design fixes it (\S\ref{sec:scoring}, Call order). \ttb{} applies action matching to selected telecom tasks, and TRAJECT-Bench
uses a judge model in part; the remaining
rows are scored as described in \S\ref{sec:related}.}
\label{tab:taucompare}
\end{table}

\section{Tool Catalogue}
\label{app:catalog}
\begin{table}[H]
\centering\scriptsize\setlength{\tabcolsep}{2pt}
\begin{tabular}{@{}lccl@{}}
\toprule
Entity domain & Tools & Writes & Example tool \\
\midrule
Knowledge \& discovery & 2 & 0 & \texttt{search\_knowledge\_base} \\
Accounts               & 3 & 0 & \texttt{get\_account\_details} \\
Deposits (FD)          & 7 & 3 & \texttt{create\_fd} \\
Deposits (RD)          & 3 & 1 & \texttt{calculate\_rd\_maturity} \\
Loans                  & 3 & 0 & \texttt{calculate\_emi} \\
Gold                   & 2 & 0 & \texttt{calculate\_gold\_loan\_ltv} \\
Cards                  & 4 & 3 & \texttt{block\_card} \\
Mandates               & 2 & 1 & \texttt{cancel\_mandate} \\
Cheques                & 2 & 1\,+\,1 & \texttt{stop\_cheque\_payment} \\
Statements \& service  & 4 & 2\,+\,1 & \texttt{raise\_request} \\
Insurance              & 1 & 0 & \texttt{get\_insurance\_details} \\
\midrule
Total                  & 33 & 11\,+\,2 & \\
\bottomrule
\end{tabular}
\caption{The 33 servicing tools by entity domain, with the number of writes.
The first number counts tools that change records; the number after the plus
sign counts tools that request a cheque book or a duplicate statement without
changing any record. The scorer treats both as writes, 13 in all. All other tools
are reads. A 34th tool transfers the customer to a human staff member.
Recurring deposits (RD) share the closure and renewal tools of fixed deposits
(FD), which are counted once, under FD.}
\label{tab:catalog}
\end{table}

\section{Example Task}
\label{app:example}
Figure~\ref{fig:exampletask} reproduces task \texttt{rl\_tlfdmat\_0197} as
released in \texttt{tasks.json}. The simulated customer is a senior citizen who prefers explanations without
jargon. The customer requests a fixed deposit (FD) of \rupee 2{,}00{,}000 for
12 months, but only after reviewing the maturity value projected from the
current rate card, and then asks to see the savings balance after the debit. The customer's instructions are written for the
simulator and are never shown to the assistant, which must obtain each detail
through the conversation and the tools. The gold chain lists five calls; for two of them,
\texttt{compare\_args} restricts argument matching to the named fields.
Three of the four distinct calls are reads: the account balance, the FD rate
card, and the maturity projection at the quoted 7.3\% rate. The final balance
read repeats the first, so a single balance read satisfies both
(\S\ref{sec:scoring}). Since \texttt{create\_fd} is the only write, the final-state check observes
only its effects: the new deposit and the debit from the savings account. Only
the action check can therefore determine whether the assistant consulted the
rate card and projected the maturity value.
\begin{figure}[t]
\centering
\begin{minipage}{\columnwidth}
\scriptsize
\begin{verbatim}
{
 "id": "rl_tlfdmat_0197",
 "description": { "purpose": "Balance -> rate ->
    maturity calc -> create FD -> balance ...", ... },
 "user_scenario": {
  "persona": "You are a senior citizen who prefers
    clear, patient explanations ...",
  "instructions": {
   "domain": "banking",
   "reason_for_call": "You want a new FD of Rs.200000
     for 12 months, but only after seeing the
     projected maturity value ...",
   "known_info": "... Amount 200000; tenure 12
     months; source savings SB5675870289.",
   "unknown_info": "Your available balance, today's
     FD rate, ...",
   "task_instructions": "Step 1: ask for your savings
     balance ... Step 5: finally ask for the savings
     balance AGAIN ..." } },
 "initial_state": { ... },
 "evaluation_criteria": {
  "actions": [
   { "name": "get_account_balance",
     "arguments": { "account_ids": ["SB5675870289"] } },
   { "name": "get_deposit_loan_rates",
     "arguments": { "product_type": "fd" } },
   { "name": "calculate_fd_maturity",
     "arguments": { "principal_amount": 200000,
                    "rate": 7.3, "tenure_months": 12 } },
   { "name": "create_fd",
     "arguments": { "principal_amount": 200000,
                    "tenure_months": 12,
                    "source_account": "SB5675870289" } },
   { "name": "get_account_balance",
     "arguments": { "account_ids": ["SB5675870289"] } } ],
  "communicate_info": [],
  "nl_assertions": [ "Agent checks funds and rate
     before projecting ..." ],
  "reward_basis": ["ACTION", "DB"] }
}
\end{verbatim}
\end{minipage}
\caption{The released task \texttt{rl\_tlfdmat\_0197} from \texttt{tasks.json},
abridged: ``\texttt{...}'' marks elided text, action identifiers and
\texttt{compare\_args} are omitted, and the rupee sign and arrows are rendered
as \texttt{Rs.} and \texttt{->}. The five actions form the gold chain; only \texttt{create\_fd}
changes the records.}
\label{fig:exampletask}
\end{figure}

\section{Dataset Statistics by Category}
\label{app:catstats}
\begin{table}[H]
\centering\small\setlength{\tabcolsep}{4pt}
\begin{tabular}{@{}llr@{}}
\toprule
Category & What it requires & Tasks \\
\midrule
Sequencing & several dependent calls & 470 \\
Edge case & a condition to recognise before acting & 266 \\
Rare tool & an infrequently used operation & 152 \\
Control & a simple request (baseline) & 112 \\
\bottomrule
\end{tabular}
\caption{The four task categories and their task counts (base split).}
\label{tab:categories}
\end{table}
\begin{table}[H]
\centering\small\setlength{\tabcolsep}{2.5pt}
\begin{tabular}{@{}lrrrrrr@{}}
\toprule
Category & Fam. & Tasks & Mean & Short\,/\,long & Reads & Write \\
\midrule
Sequencing & 20 & 470 & 3.64 & 279\,/\,191 & 63.1\% & 449 \\
Edge case  & 14 & 266 & 4.01 & 93\,/\,173  & 69.3\% & 266 \\
Rare tool  & 8  & 152 & 4.01 & 39\,/\,113  & 75.0\% & 152 \\
Control    & 8  & 112 & 1.38 & 112\,/\,0   & 87.7\% & 19 \\
\bottomrule
\end{tabular}
\caption{Composition of the four task categories in the released base split.
\textbf{Fam.}: task families (Table~\ref{tab:families}). \textbf{Mean}: mean number of calls in the gold
chain (graded categories 3--6 calls, control 1--3). \textbf{Short\,/\,long}: tasks
with at most three and with four or more gold-chain calls. \textbf{Reads}: share of
gold-chain calls that are reads. \textbf{Write}: tasks whose gold chain contains
at least one write.}
\label{tab:catstats}
\end{table}

\begin{table*}[t]
\centering\scriptsize\setlength{\tabcolsep}{2.5pt}
\begin{minipage}[t]{0.485\textwidth}\centering
\begin{tabular}[t]{@{}lr>{\raggedright\arraybackslash}p{4.9cm}@{}}
\toprule
Family & Tasks & Gold chain \\
\midrule
\multicolumn{3}{@{}l}{\emph{Sequencing} (20 families, 470 tasks)}\\
\texttt{sqfd} & 29 & Balance $\rightarrow$ rates $\rightarrow$ book FD $\rightarrow$ balance \\
\texttt{sqclose} & 29 & FD details $\rightarrow$ closure quote $\rightarrow$ close $\rightarrow$ balance \\
\texttt{sqrenew} & 29 & FD details $\rightarrow$ change renewal $\rightarrow$ verify \\
\texttt{sqkb} & 27 & Policy $\rightarrow$ closure quote $\rightarrow$ change renewal \\
\texttt{sqfrz} & 30 & Card status $\rightarrow$ freeze $\rightarrow$ verify \\
\texttt{sqctl} & 29 & Card status $\rightarrow$ set controls $\rightarrow$ verify \\
\texttt{sqdisp} & 27 & Transactions $\rightarrow$ block card $\rightarrow$ dispute \\
\texttt{sqmand} & 30 & List mandates $\rightarrow$ cancel $\rightarrow$ list \\
\texttt{sqcheq} & 25 & Transactions $\rightarrow$ stop cheque $\rightarrow$ cheque book \\
\texttt{sqreloc} & 25 & New address $\rightarrow$ cheque book $\rightarrow$ statement \\
\texttt{sq2cardblk} & 21 & Card status $\rightarrow$ block $\rightarrow$ replacement $\rightarrow$ verify \\
\texttt{sq2emi} & 21 & Loan $\rightarrow$ rate $\rightarrow$ EMI $\rightarrow$ eligibility policy \\
\texttt{sq2fdladder} & 21 & FD details $\rightarrow$ quote $\rightarrow$ close $\rightarrow$ rate $\rightarrow$ new FD $\rightarrow$ balance \\
\texttt{sq2gold} & 23 & Gold rate $\rightarrow$ LTV $\rightarrow$ policy $\rightarrow$ loan request \\
\texttt{sq2insur} & 15 & Policy details $\rightarrow$ claim rules $\rightarrow$ claim request \\
\texttt{sq2loanfc} & 19 & Loan $\rightarrow$ foreclosure quote $\rightarrow$ request \\
\texttt{sq2mandaudit} & 15 & List $\rightarrow$ cancel $\rightarrow$ list $\rightarrow$ cancel $\rightarrow$ list \\
\texttt{sq2rdsize} & 21 & Balance $\rightarrow$ rate $\rightarrow$ RD maturity $\rightarrow$ penalty rule $\rightarrow$ book RD $\rightarrow$ balance \\
\texttt{sq2seniorfd} & 11 & Balance $\rightarrow$ senior benefit $\rightarrow$ rate $\rightarrow$ book FD $\rightarrow$ balance \\
\texttt{sq2statement} & 23 & Transactions $\rightarrow$ dispute $\rightarrow$ statement \\
\midrule
\multicolumn{3}{@{}l}{\emph{Control} (8 families, 112 tasks)}\\
\texttt{hpbal} & 14 & Balance enquiry \\
\texttt{hpcard} & 11 & Card details \\
\texttt{hpemi} & 13 & Loan rate $\rightarrow$ EMI \\
\texttt{hpgold} & 10 & Gold rate $\rightarrow$ LTV $\rightarrow$ offers \\
\texttt{hpkb} & 12 & General policy question \\
\texttt{hprate} & 11 & RD rate enquiry \\
\texttt{hpservice} & 27 & Service request (cheque book, statement, product or insurance query) \\
\texttt{hptxn} & 14 & Recent transactions \\
\bottomrule
\end{tabular}
\end{minipage}\hfill
\begin{minipage}[t]{0.485\textwidth}\centering
\begin{tabular}[t]{@{}lr>{\raggedright\arraybackslash}p{4.9cm}@{}}
\toprule
Family & Tasks & Gold chain or condition \\
\midrule
\multicolumn{3}{@{}l}{\emph{Edge case} (14 families, 266 tasks)}\\
\texttt{edgablk2x} & 19 & Block a card; decline a second block \\
\texttt{edgaineligible} & 19 & Senior rate asked by an ineligible customer \\
\texttt{edgadestructive} & 19 & Quote and confirmation before closing an RD \\
\texttt{edgapolicy} & 19 & Dispute without disclosing internal codes \\
\texttt{edgamandwrong} & 19 & Mandate named by payee, wrong ID given \\
\texttt{edgaphantom} & 19 & Account that is not on record \\
\texttt{edgbcarddisamb} & 19 & Two cards; freeze the one named \\
\texttt{edgbstmtwin} & 17 & Statement period before account opening \\
\texttt{edgbmindchange} & 19 & FD tenure changed during the request \\
\texttt{edgbtwocards} & 21 & Freeze one card; block and replace the other \\
\texttt{edgbmatelig} & 19 & Cancel only the active mandate \\
\texttt{edgainsuff} & 19 & FD larger than the balance \\
\texttt{edgbfdcap} & 21 & FD limited by the minimum-balance rule \\
\texttt{edgbloanrd} & 17 & RD sized around an existing loan \\
\midrule
\multicolumn{3}{@{}l}{\emph{Rare tool} (8 families, 152 tasks)}\\
\texttt{tlacctdet} & 20 & Balance $\rightarrow$ account details $\rightarrow$ statement \\
\texttt{tlfdmat} & 21 & Balance $\rightarrow$ rate $\rightarrow$ FD maturity $\rightarrow$ book FD $\rightarrow$ balance \\
\texttt{tlrdmat} & 19 & Balance $\rightarrow$ rate $\rightarrow$ RD maturity $\rightarrow$ book RD $\rightarrow$ balance \\
\texttt{tlgoldltv} & 19 & Gold rate $\rightarrow$ LTV $\rightarrow$ loan offers $\rightarrow$ request \\
\texttt{tlinsur} & 14 & Balance $\rightarrow$ policy $\rightarrow$ payment modes $\rightarrow$ auto-debit request \\
\texttt{tlprodoff} & 21 & Balance $\rightarrow$ offers $\rightarrow$ rate $\rightarrow$ record interest \\
\texttt{tlrddet} & 19 & Balance $\rightarrow$ RD details $\rightarrow$ closure quote $\rightarrow$ change renewal \\
\texttt{tlreqstat} & 19 & Balance $\rightarrow$ request status $\rightarrow$ follow-up request \\
\bottomrule
\end{tabular}
\end{minipage}
\caption{The 50 task families. Family identifiers appear in each task
identifier (e.g.\ \texttt{rl\_tlfdmat\_0197}). For sequencing, rare-tool, and
control families, the last column gives the typical gold chain in call
order; for edge-case families, it gives the condition that the assistant must
handle. Edge-case families are listed in the four groups of Appendix~\ref{app:catstats}.}
\label{tab:families}
\end{table*}

\noindent Table~\ref{tab:families} lists all 50 families. The 14 edge-case families fall into four groups, according to the
condition that the assistant must handle. The first group (4 families, 76
tasks) requires declining an invalid or unsafe request, such as repeating an
irreversible card block, applying a benefit for which the customer is
ineligible, skipping confirmation under pressure, or disclosing internal codes.
The second (5 families, 93 tasks) requires correcting or clarifying information
supplied by the customer, such as a misremembered mandate number, an account
that is not on record, an unspecified card, a statement period that precedes
the account's opening, or a changed tenure. The third (2 families, 40 tasks)
requires applying the correct action to each of several similar cards or
mandates. The fourth (3 families, 57 tasks) requires limiting a commitment to
what the account supports. No gold chain includes a transfer to a human
staff member. 

{\raggedright Each rare-tool family centres on one of eight tools (\texttt{get\_account\_details},
\texttt{calculate\_fd\_maturity}, \texttt{calculate\_rd\_maturity},
\texttt{calculate\_gold\_loan\_ltv}, \texttt{get\_insurance\_details},
\texttt{get\_products\_and\_offers}, \texttt{get\_rd\_details},
\texttt{get\_request\_status}). Each of these tools appears in at least 19
gold chains, and \texttt{calculate\_fd\_maturity} and
\texttt{get\_request\_status} appear only in rare-tool tasks.\par}

\section{Scoring Pseudocode}
\label{app:pseudocode}
Algorithm~\ref{alg:score} gives the scoring procedure for one session.

\begin{algorithm}[H]
\small
\caption{Session score.}
\label{alg:score}
\begin{algorithmic}[1]
\State \textbf{input:} session $\sigma$, task $t$, tool types $T$
\If{$\sigma.\text{termination} \notin \{\textsc{agent\_stop}, \textsc{user\_stop}\}$}
  \State \Return $0$ \Comment{session cut off before a normal end}
\EndIf
\State $P \gets$ tool calls in $\sigma$
\State $G \gets t.\text{gold\_actions}$
\Statex \textbf{Action check (membership + write purity):}
\State $a \gets \big[\, \forall g \in G:\ \exists p \in P:\ \textsc{match}(g,p) \,\big]$
\ForAll{$p \in P$ with $T[p]=\textsc{write}$}
  \If{$\neg\, \exists g \in G:\ T[g]=\textsc{write} \wedge \textsc{match}(g,p)$}
    \State $a \gets \textsc{false}$ \Comment{unrequested write}
  \EndIf
\EndFor
\Statex \textbf{Final-state check (oracle replay):}
\State $s^\star \gets \textsc{replay}(G,\ t.\text{seed})$
\State $d \gets \big[\, \textsc{hash}(\textsc{norm}(\sigma.\text{state})) = \textsc{hash}(\textsc{norm}(s^\star)) \,\big]$
\State \Return $\mathbf{1}[a] \times \mathbf{1}[d]$
\end{algorithmic}
\end{algorithm}

\noindent \textsc{match} compares the tool name and the arguments specified by the
gold-chain call under the rules of \S\ref{sec:scoring}, ignoring fields that vary between runs; \textsc{norm} removes
those fields and ignores the position of entries in collections keyed by
generated identifiers. \textsc{write} includes the two tools that request a cheque
book or a duplicate statement, and a transfer to a human staff member ends the session with \textsc{user\_stop}.

\section{Scorer Audit}
\label{app:audit}
\begin{table}[H]
\centering\small\setlength{\tabcolsep}{3pt}
\begin{tabular}{@{}lc>{\raggedright\arraybackslash}p{0.50\columnwidth}@{}}
\toprule
Check & In score & Agreement \\
\midrule
Action + final state & yes & same verdict on all 600 sessions; all 705 paired calls match \\
Communication & no & all 400 sampled tasks pass \\
Judge assertion & no & the judge agrees on 87.5\% and gives less credit in every disagreement \\
\bottomrule
\end{tabular}
\caption{Agreement of the scorer with an independent re-implementation.
\textbf{In score}: whether the check affects the released score. The
communication check only flags replies that expose tool names or internal
codes; it does not affect the score.}
\label{tab:audit}
\end{table}

\section{Additional Results}
\label{app:exp34}
\begin{figure}[H]
\centering
\ifprebuiltfigs
\includegraphics{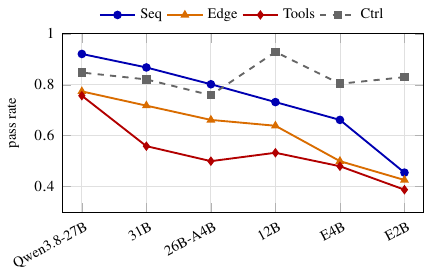}
\else
\begin{tikzpicture}
\begin{axis}[
  width=\columnwidth, height=4.6cm,
  xtick=data, symbolic x coords={Qwen3.8-27B,31B,26B-A4B,12B,E4B,E2B},
  x tick label style={font=\scriptsize, rotate=28, anchor=north east, inner sep=1pt, yshift=-3pt},
  ymin=0.30, ymax=1.0, ylabel={pass rate}, ylabel near ticks,
  y tick label style={font=\scriptsize}, ylabel style={font=\scriptsize},
  legend style={font=\scriptsize, at={(0.5,1.02)}, anchor=south, draw=none, fill=none},
  legend columns=4, grid=major, grid style={black!12}, tick style={black!30},
  every axis plot/.append style={line width=0.9pt, mark size=1.6pt}, legend cell align=left, enlarge x limits=0.06]
\addplot+[color=blue!70!black, mark=*, mark options={solid, fill=blue!70!black}] coordinates
 {(E2B,0.455)(E4B,0.662)(12B,0.732)(26B-A4B,0.802)(31B,0.868)(Qwen3.8-27B,0.921)};
\addlegendentry{Seq}
\addplot+[color=orange!85!black, mark=triangle*, mark options={solid, fill=orange!85!black}] coordinates
 {(E2B,0.426)(E4B,0.500)(12B,0.639)(26B-A4B,0.662)(31B,0.718)(Qwen3.8-27B,0.774)};
\addlegendentry{Edge}
\addplot+[color=red!70!black, mark=diamond*, mark options={solid, fill=red!70!black}] coordinates
 {(E2B,0.388)(E4B,0.480)(12B,0.533)(26B-A4B,0.500)(31B,0.559)(Qwen3.8-27B,0.757)};
\addlegendentry{Tools}
\addplot+[color=black!60, mark=square*, mark options={solid, fill=black!60}, dashed] coordinates
 {(E2B,0.830)(E4B,0.804)(12B,0.929)(26B-A4B,0.759)(31B,0.821)(Qwen3.8-27B,0.848)};
\addlegendentry{Ctrl}
\end{axis}
\end{tikzpicture}
\fi
\caption{Pass rate by category for the six models, sorted by overall pass
rate (Table~\ref{tab:ladder}). Among the four dense Gemma models, pass rates in
each graded category rise with model size; control pass rates do not.}
\label{fig:cat}
\end{figure}
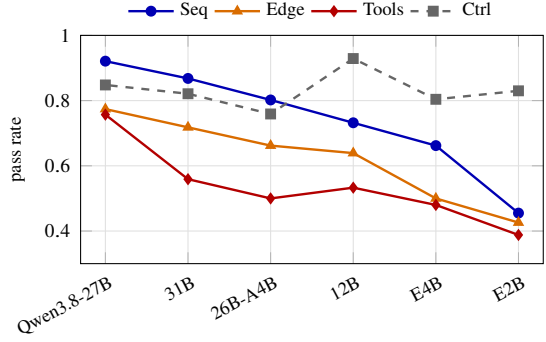

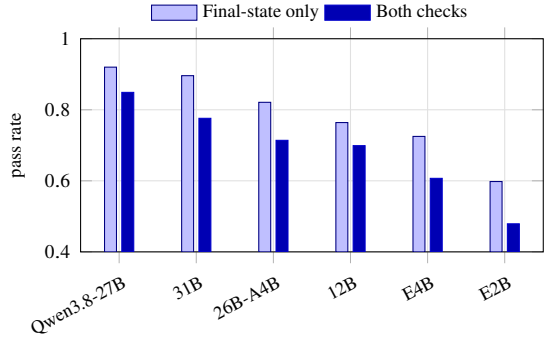
\begin{figure}[H]
\centering
\ifprebuiltfigs
\includegraphics{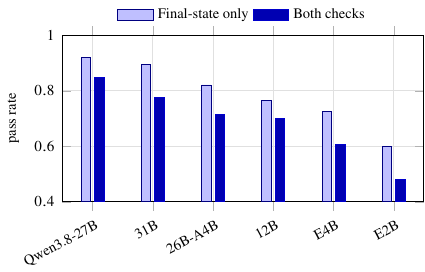}
\else
\begin{tikzpicture}
\begin{axis}[
  width=\columnwidth, height=4.4cm, ybar, bar width=4.5pt,
  symbolic x coords={Qwen3.8-27B,31B,26B-A4B,12B,E4B,E2B}, xtick=data,
  x tick label style={font=\scriptsize, rotate=28, anchor=north east, inner sep=1pt, yshift=-3pt},
  ymin=0.40, ymax=1.0, ylabel={pass rate}, ylabel near ticks,
  y tick label style={font=\scriptsize}, ylabel style={font=\scriptsize},
  legend style={font=\scriptsize, at={(0.5,1.03)}, anchor=south, draw=none, fill=none, legend columns=2},
  area legend, legend cell align=left,
  grid=major, grid style={black!12}, tick style={black!30},
  enlarge x limits=0.10]
\addplot+[fill=blue!25, draw=blue!50!black] coordinates
 {(Qwen3.8-27B,0.920)(31B,0.896)(26B-A4B,0.821)(12B,0.764)(E4B,0.725)(E2B,0.598)};
\addlegendentry{Final-state only}
\addplot+[fill=blue!70!black, draw=blue!80!black] coordinates
 {(Qwen3.8-27B,0.849)(31B,0.776)(26B-A4B,0.714)(12B,0.699)(E4B,0.607)(E2B,0.479)};
\addlegendentry{Both checks}
\end{axis}
\end{tikzpicture}
\fi
\caption{Pass rate with the final-state check alone (light bars) and with both
checks (dark bars), computed on the same sessions (Table~\ref{tab:exp1}).
The difference between the bars is the share of procedural violations.}
\label{fig:inflate}
\end{figure}

Figure~\ref{fig:inflate} plots the pass rates of Table~\ref{tab:exp1},
Figure~\ref{fig:cat} the per-category pass rates of Table~\ref{tab:ladder}, and
Table~\ref{tab:causes} gives the causes of action-check failure discussed in
\S\ref{sec:exp1}.

\begin{table}[H]
\centering\small\setlength{\tabcolsep}{4pt}
\begin{tabular}{@{}lrrr@{}}
\toprule
                & Missing & Missing & Unrequested \\
Assistant model & read & write & write \\
\midrule
Qwen3.8-27B      & 48  & 84  & 63 \\
Gemma\,4 31B     & 107 & 127 & 73 \\
Gemma\,4 26B-A4B & 143 & 170 & 73 \\
Gemma\,4 12B     & 81  & 142 & 54 \\
Gemma\,4 E4B     & 178 & 160 & 48 \\
Gemma\,4 E2B     & 195 & 186 & 97 \\
\bottomrule
\end{tabular}
\caption{Causes of action-check failure, counted over all failed sessions: a
missing required read, a missing required write, or an unrequested write
detected by write purity. A session can have more than one cause, so a row may exceed the number of failed sessions.}
\label{tab:causes}
\end{table}

\subsection{Violations by service area}
\label{sec:exp6}
Pooled over the six models, $20.8\%$ of sessions on loan tasks are procedural
violations (\S\ref{sec:exp1}), compared with $7.9$--$10.5\%$ in each of the
other three service areas (Table~\ref{tab:usecase}).

\paragraph{Service areas.} Each task is assigned to one area: deposits, loans (including gold loans), or
cards. The area is the one whose tools account for most of the area-tool calls
in the task's gold chain, and ties go to the area listed first. The shared
rate-card lookup (\texttt{get\_deposit\_loan\_rates}) is not assigned to any
area. Tasks that use none of the area tools are classified as everyday banking.

\begin{table}[H]
\centering\small\setlength{\tabcolsep}{1.8pt}
\begin{tabular}{@{}lrrrrr@{}}
\toprule
Assistant model & Everyday & Deposits & Loans & Cards & Total \\
\midrule
Qwen3.8-27B      & 13 & 22 & 22 & 14 & 71 \\
Gemma\,4 31B     & 45 & 31 & 29 & 15 & 120 \\
Gemma\,4 26B-A4B & 53 & 12 & 23 & 19 & 107 \\
Gemma\,4 12B     & 14 & 19 & 15 & 17 & 65 \\
Gemma\,4 E4B     & 36 & 39 & 22 & 21 & 118 \\
Gemma\,4 E2B     & 34 & 38 & 20 & 26 & 118 \\
\midrule
Pooled           & 195 & 161 & 131 & 112 & 599 \\
Rate (\%)        & 8.6 & 7.9 & 20.8 & 10.5 & -- \\
\bottomrule
\end{tabular}
\caption{Procedural violations by model and service area. \textbf{Rate}: each
pooled count divided by the number of sessions in that area, that is, six
models times the number of tasks (378 everyday, 340 deposit, 105 loan and 177 card tasks).}
\label{tab:usecase}
\end{table}

\subsection{Pass rates by gold-chain length}
\label{app:len}
Figure~\ref{fig:len} and Table~\ref{tab:exp3} give the pass rate for each
gold-chain length. E2B scores lowest on five-call tasks, whereas every other
model scores higher on five- and six-call tasks than on four-call tasks; the gap
between Qwen3.8-27B and E2B narrows to $0.423$ on six-call tasks.
Table~\ref{tab:compound} divides each graded category into short tasks (at most
three calls) and long tasks (four or more). Qwen3.8-27B scores higher on long
sequencing and edge-case tasks than on short ones. E2B falls from $0.543$ to $0.364$ on edge cases.
\begin{table}[H]
\centering\small\setlength{\tabcolsep}{2pt}
\begin{tabular}{@{}lrrrrrr@{}}
\toprule
Assistant model & L1 & L2 & L3 & L4 & L5 & L6 \\
\midrule
Qwen3.8-27B      & 0.928 & 0.933 & 0.849 & 0.763 & 0.960 & 0.898 \\
Gemma\,4 31B     & 0.928 & 0.867 & 0.812 & 0.625 & 0.843 & 0.881 \\
Gemma\,4 26B-A4B & 0.867 & 0.800 & 0.725 & 0.622 & 0.709 & 0.864 \\
Gemma\,4 12B     & 1.000 & 0.867 & 0.715 & 0.564 & 0.701 & 0.780 \\
Gemma\,4 E4B     & 0.880 & 0.867 & 0.706 & 0.419 & 0.504 & 0.593 \\
Gemma\,4 E2B     & 0.904 & 0.933 & 0.502 & 0.361 & 0.346 & 0.475 \\
\midrule
$n$ tasks        & 83 & 15 & 425 & 291 & 127 & 59 \\
\bottomrule
\end{tabular}
\caption{Pass rate by number of calls in the gold chain.}
\label{tab:exp3}
\end{table}
\begin{table}[H]
\centering\small\setlength{\tabcolsep}{2pt}
\begin{tabular}{@{}lrrrrrr@{}}
\toprule
 & \multicolumn{2}{c}{Seq} & \multicolumn{2}{c}{Edge} & \multicolumn{2}{c}{Tools} \\
\cmidrule(lr){2-3}\cmidrule(lr){4-5}\cmidrule(l){6-7}
Assistant model & Short & Long & Short & Long & Short & Long \\
\midrule
Qwen3.8-27B      & 0.907 & 0.942 & 0.731 & 0.798 & 0.923 & 0.699 \\
Gemma\,4 31B     & 0.903 & 0.817 & 0.753 & 0.699 & 0.538 & 0.566 \\
Gemma\,4 26B-A4B & 0.817 & 0.780 & 0.667 & 0.659 & 0.436 & 0.522 \\
Gemma\,4 12B     & 0.742 & 0.717 & 0.742 & 0.584 & 0.513 & 0.540 \\
Gemma\,4 E4B     & 0.810 & 0.445 & 0.516 & 0.491 & 0.564 & 0.451 \\
Gemma\,4 E2B     & 0.498 & 0.393 & 0.543 & 0.364 & 0.513 & 0.345 \\
\bottomrule
\end{tabular}
\caption{Pass rate on short (at most three gold-chain calls) and long (four
or more) tasks within each graded category: sequencing (\textbf{Seq}), edge case
(\textbf{Edge}) and rare tool (\textbf{Tools}). Task counts are given in Table~\ref{tab:catstats}.}
\label{tab:compound}
\end{table}
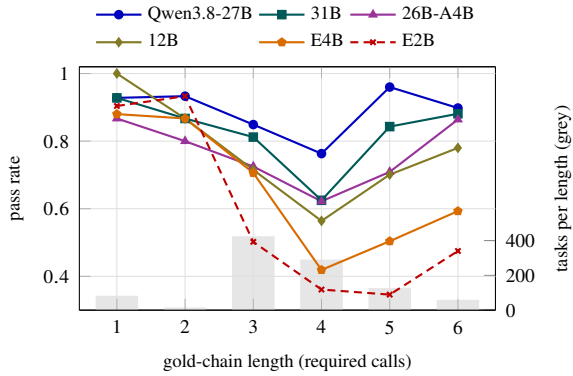
\begin{figure}[H]
\centering
\ifprebuiltfigs
\includegraphics{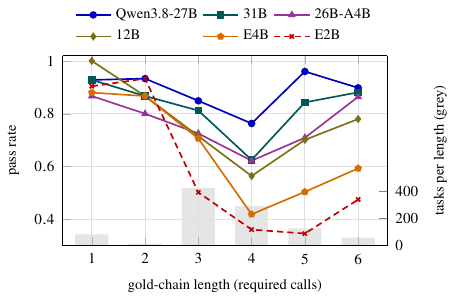}
\else
\begin{tikzpicture}
\begin{axis}[
  width=0.92\columnwidth, height=4.8cm, xmin=0.45, xmax=6.55,
  axis y line*=right, ymin=0, ymax=1400,
  ylabel={tasks per length (grey)}, ylabel style={font=\scriptsize},
  ytick={0,200,400}, y tick label style={font=\scriptsize},
  xtick=\empty]
\addplot[ybar, bar width=16pt, fill=black!10, draw=none] coordinates
 {(1,83)(2,15)(3,425)(4,291)(5,127)(6,59)};
\end{axis}
\begin{axis}[
  width=0.92\columnwidth, height=4.8cm, xmin=0.45, xmax=6.55,
  axis y line*=left,
  xtick={1,2,3,4,5,6}, xlabel={gold-chain length (required calls)},
  xlabel style={font=\scriptsize}, x tick label style={font=\scriptsize},
  ymin=0.30, ymax=1.02, ylabel={pass rate}, ylabel near ticks,
  y tick label style={font=\scriptsize}, ylabel style={font=\scriptsize},
  legend style={font=\scriptsize, at={(0.5,1.02)}, anchor=south, draw=none, fill=none},
  legend columns=3, grid=major, grid style={black!12}, tick style={black!30},
  every axis plot/.append style={line width=0.8pt, mark size=1.4pt}, legend cell align=left]
\addplot+[color=blue!75!black, mark=*, mark options={solid, fill=blue!75!black}] coordinates {(1,0.928)(2,0.933)(3,0.849)(4,0.763)(5,0.960)(6,0.898)}; \addlegendentry{Qwen3.8-27B}
\addplot+[color=teal!70!black, mark=square*, mark options={solid, fill=teal!70!black}] coordinates {(1,0.928)(2,0.867)(3,0.812)(4,0.625)(5,0.843)(6,0.881)}; \addlegendentry{31B}
\addplot+[color=violet!80, mark=triangle*, mark options={solid, fill=violet!80}] coordinates {(1,0.867)(2,0.800)(3,0.725)(4,0.622)(5,0.709)(6,0.864)}; \addlegendentry{26B-A4B}
\addplot+[color=olive!80!black, mark=diamond*, mark options={solid, fill=olive!80!black}] coordinates {(1,1.000)(2,0.867)(3,0.715)(4,0.564)(5,0.701)(6,0.780)}; \addlegendentry{12B}
\addplot+[color=orange!85!black, mark=pentagon*, mark options={solid, fill=orange!85!black}] coordinates {(1,0.880)(2,0.867)(3,0.706)(4,0.419)(5,0.504)(6,0.593)}; \addlegendentry{E4B}
\addplot+[color=red!75!black, mark=x, mark options={solid, fill=red!75!black}] coordinates {(1,0.904)(2,0.933)(3,0.502)(4,0.361)(5,0.346)(6,0.475)}; \addlegendentry{E2B}
\end{axis}
\end{tikzpicture}
\fi
\caption{Pass rate by gold-chain length (Table~\ref{tab:exp3}); grey
bars give the number of tasks of each length (right axis).}
\label{fig:len}
\end{figure}

\subsection{Failure types}
\label{sec:exp4}
\begin{figure}[H]
\centering
\ifprebuiltfigs
\includegraphics{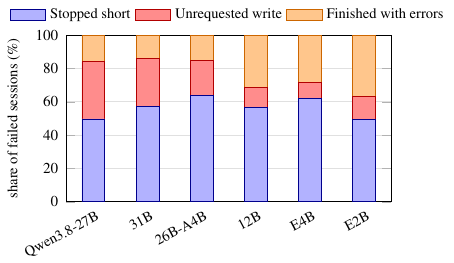}
\else
\begin{tikzpicture}
\begin{axis}[
  width=0.92\columnwidth, height=4.4cm, ybar stacked, bar width=11pt,
  symbolic x coords={Qwen3.8-27B,31B,26B-A4B,12B,E4B,E2B}, xtick=data,
  x tick label style={font=\scriptsize, rotate=28, anchor=north east, inner sep=1pt, yshift=-3pt},
  ymin=0, ymax=100, ylabel={share of failed sessions (\%)}, ylabel near ticks,
  y tick label style={font=\scriptsize}, ylabel style={font=\scriptsize},
  legend style={font=\scriptsize, at={(0.5,1.03)}, anchor=south, draw=none, fill=none, legend columns=3},
  area legend,
  grid=major, grid style={black!12}, tick style={black!30},
  enlarge x limits=0.10]
\addplot+[fill=blue!30, draw=blue!55!black] coordinates
 {(Qwen3.8-27B,49.7)(31B,57.1)(26B-A4B,64.0)(12B,56.5)(E4B,62.1)(E2B,49.6)};
\addlegendentry{Stopped short}
\addplot+[fill=red!45, draw=red!70!black] coordinates
 {(Qwen3.8-27B,34.4)(31B,29.0)(26B-A4B,21.0)(12B,12.3)(E4B,9.7)(E2B,13.5)};
\addlegendentry{Unrequested write}
\addplot+[fill=orange!45, draw=orange!80!black] coordinates
 {(Qwen3.8-27B,15.9)(31B,13.8)(26B-A4B,15.0)(12B,31.2)(E4B,28.2)(E2B,36.9)};
\addlegendentry{Finished with errors}
\end{axis}
\end{tikzpicture}
\fi
\caption{Distribution of failure types among each model's failed sessions
(Table~\ref{tab:exp4}). Stronger models tend to have a larger share of
unrequested writes and a smaller share of sessions finished with errors.}
\label{fig:modes}
\end{figure}
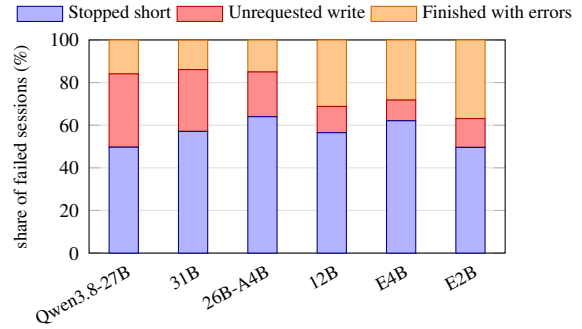

We measure the fraction of gold-chain calls that each session made and assign
every failed session to one of three types (Table~\ref{tab:exp4} and
Figure~\ref{fig:modes}). A session that \emph{stopped short} ended or was
handed off before the procedure was complete. A session with an
\emph{unrequested write} made a change that the task did not request. A session
that \emph{finished with errors} called every tool in the gold chain but
used incorrect arguments in at least one call, or ended with incorrect records.

Stopping short accounts for $49.6$--$64.0\%$ of failed sessions. The other two
types vary in opposite directions. Unrequested writes account for $34.4\%$ of failures for Qwen3.8-27B and
$29.0\%$ for the 31B, but for only $9.7$--$13.5\%$ for the three weakest
models. Finishing with errors accounts for $13.8$--$15.9\%$ for the three
strongest models and $28.2$--$36.9\%$ for the three weakest. On average, the
strongest model completes $0.911$ of its gold chain and the weakest $0.763$.
\begin{table}[H]
\centering\small\setlength{\tabcolsep}{3pt}
\begin{tabular}{@{}lrrrr@{}}
\toprule
                & Compl. & Stopped & Unreq. & Finished \\
Assistant model & frac.  & short & write & w.\ errors \\
\midrule
Qwen3.8-27B      & 0.911 & 49.7 & 34.4 & 15.9 \\
Gemma\,4 31B     & 0.850 & 57.1 & 29.0 & 13.8 \\
Gemma\,4 26B-A4B & 0.811 & 64.0 & 21.0 & 15.0 \\
Gemma\,4 12B     & 0.863 & 56.5 & 12.3 & 31.2 \\
Gemma\,4 E4B     & 0.779 & 62.1 & \phantom{0}9.7 & 28.2 \\
Gemma\,4 E2B     & 0.763 & 49.6 & 13.5 & 36.9 \\
\bottomrule
\end{tabular}
\caption{Failure types. \textbf{Compl.\ frac.}: mean fraction of the gold chain
completed, over all sessions. The last three columns give the share (\%) of
\emph{failed} sessions that stopped short, made an unrequested write, or
finished with errors.}
\label{tab:exp4}
\end{table}

% Column break: balances the two columns of the last page.

\subsection{Outcomes of two attempts}
\label{app:band}
Every other reported result derives from a single attempt per task. To assess
consistency, we ran the base split a second time for the five Gemma models;
Table~\ref{tab:band} groups the tasks by the pair of outcomes.
\begin{table}[H]
\centering\small\setlength{\tabcolsep}{4pt}
\begin{tabular}{@{}lrrr@{}}
\toprule
Assistant model & Fail both & Mixed & Pass both \\
\midrule
Gemma\,4 31B     & 16.6 & 14.3 & 69.1 \\
Gemma\,4 26B-A4B & 18.6 & 20.3 & 61.1 \\
Gemma\,4 12B     & 18.3 & 22.7 & 59.1 \\
Gemma\,4 E4B     & 27.5 & 26.0 & 46.5 \\
Gemma\,4 E2B     & 39.9 & 23.1 & 36.9 \\
\bottomrule
\end{tabular}
\caption{Share of tasks (\%) by the outcomes of two attempts (\textbf{Mixed}:
one attempt passes and one fails). The share passing both attempts rises with model strength.}
\label{tab:band}
\end{table}

\end{document}